\documentclass[letterpaper,10pt,journal,twoside]{IEEEtran}
\IEEEoverridecommandlockouts

\usepackage{cite}
\usepackage{amsmath,amssymb,amsfonts}
\usepackage{algorithmic}
\usepackage{graphicx}
\usepackage{textcomp}
\usepackage{xcolor}
\usepackage{multirow}
\usepackage{tabularx}
\usepackage{array}
\usepackage{float}
\newcolumntype{Y}{>{\centering\arraybackslash}X}

\def\BibTeX{{\rm B\kern-.05em{\sc i\kern-.025em b}\kern-.08em
    T\kern-.1667em\lower.7ex\hbox{E}\kern-.125emX}}

\begin{document}

\title{TRACE: Learned Proprioceptive Odometry for Legged Robots under Unreliable Contact Conditions}

\author{Taehyeon~Kong$^{1}$, Woojin~Kim$^{1}$, and~Jemin~Hwangbo$^{1*}$%
\thanks{This work has been submitted to the IEEE for possible publication. Copyright may be transferred without notice, after which this version may no longer be accessible.

$^{1}$The authors are with the Korea Advanced Institute of Science and Technology (KAIST), Yuseong-gu, Daejeon 34141, Republic of Korea (e-mail: \{ballbug12, woojin0624, jhwangbo\}@kaist.ac.kr).
$^{*}$Jemin Hwangbo is the corresponding author.}}

\maketitle

\begin{abstract}
In this paper, we present \textbf{TRACE} (Tokenized Robust Attention for Contact-Aware Estimation), an end-to-end learned proprioceptive odometry estimator for legged robots under unreliable contact conditions. The proposed estimator directly predicts relative displacement, relative rotation, and body-frame velocity from a recent history of onboard inertial and joint measurements. To improve robustness under unreliable contact conditions, we introduce a foot-aware cross-attention module that adaptively weights IMU and leg-wise kinematic tokens without relying on manually defined contact or slip thresholds. The estimator is trained with direct supervision and two physics-inspired auxiliary losses that promote kinematic consistency and reliable use of leg information. To reduce policy-specific overfitting and consequently improve sim-to-real transfer, simulation training incorporates policy randomization, followed by partial real-world fine-tuning of the temporal encoder and prediction head. Experiments across diverse indoor and outdoor terrains demonstrate consistent reductions in position drift compared with classical filtering-based, hybrid, and purely learning-based baselines. Ablation studies further validate the contributions of the proposed training objectives, policy randomization, and real-world fine-tuning, particularly under unreliable contacts and sim-to-real mismatch.
\end{abstract}


\IEEEpeerreviewmaketitle

\section{Introduction}

\IEEEPARstart{R}{eliable} state estimation is essential for legged robots to plan and navigate complex environments~\cite{barfoot2024state}. Existing approaches often use exteroceptive sensors such as cameras, LiDAR, and GPS to provide geometric or global information~\cite{zhang2023perception}. However, these sensors can degrade under poor illumination, motion blur, textureless scenes, adverse weather, or indoor and occluded environments~\cite{bijelic2020seeing,zafari2019survey}. Proprioceptive odometry instead estimates ego-motion using onboard inertial and joint sensors~\cite{bloesch2013state}, making it useful when exteroceptive sensing is unavailable or unreliable. However, accurate proprioceptive odometry remains challenging because noisy measurements and unreliable foot--ground contacts caused by slip, compliant or deformable terrain, and foot rolling can introduce persistent estimation errors that accumulate into substantial drift over time.

Classical filtering-based estimators, especially invariant extended Kalman filter (\textbf{IEKF}) based approaches~\cite{hartley2020contact}, have become a common framework for contact-aided state estimation. These methods usually assume that stance feet have zero velocity relative to the ground. However, accurate contact state is not directly observable and must be inferred from proprioceptive measurements. Previous methods have used ground reaction forces~\cite{camurri2017probabilistic, fink2020proprioceptive} or gait phase information~\cite{piperakis2019unsupervised} to estimate contact, but they often require carefully tuned thresholds and may become inaccurate under complex terrain or dynamic locomotion. Moreover, their performance can be significantly degraded when unmodeled contact effects such as slip, impact, terrain deformation, and foot rolling violate the zero-velocity assumption. Their reliance on manually tuned covariance settings also makes it difficult to maintain robust performance across diverse terrains and locomotion conditions.

\begin{figure}[t]
    \centering
    \includegraphics[width=\columnwidth]{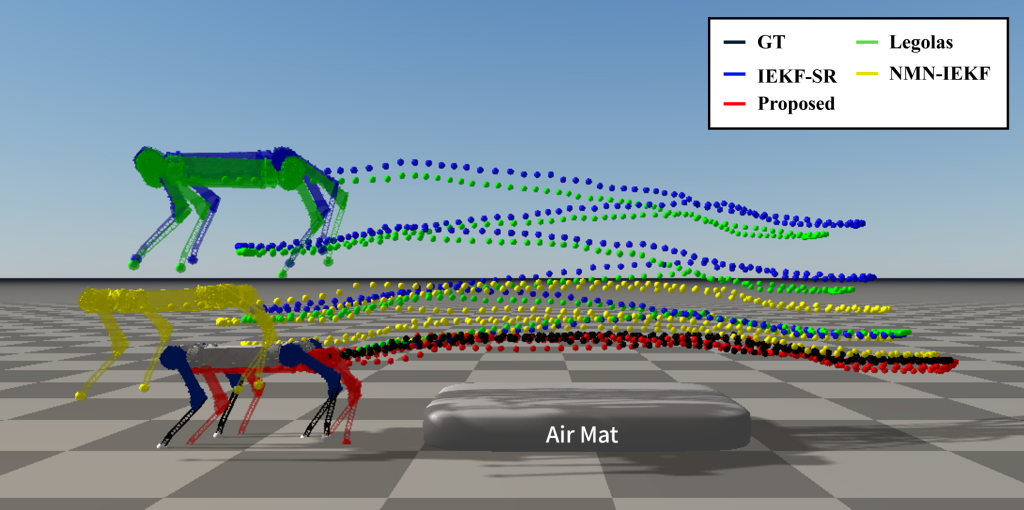}
    \caption{\textbf{Position estimation comparison on deformable terrain.}
    The robot traverses back and forth over an air-mat terrain. The proposed end-to-end learned proprioceptive odometry estimator effectively handles unreliable contacts and reduces the position ATE by approximately \textbf{53.8\%} in the soft-terrain experiment.}
    \label{fig:soft_comparison}
\end{figure}

To address these limitations, learning-based proprioceptive estimators have been studied in both hybrid and end-to-end forms. Hybrid methods augment model-based filters with neural networks: Lin et al.~\cite{lin2022legged} learned contact estimates for a contact-aided IEKF, Youm et al.~\cite{youm2025legged} proposed \textbf{NMN-IEKF} to predict foot-contact probabilities and body-frame velocity as IEKF measurements, Sun et al.~\cite{sun2025proprioceptive} introduced a CNN-based slip detector, and Lee and Kim~\cite{lee2026attention} proposed \textbf{AttenNKF} to compensate for slip-induced estimation errors. More recently, Seo et al.~\cite{seo2026gait} introduced \textbf{GAIT}, which applies attention over inertial and leg tokens to predict body-frame velocity and its uncertainty for use in an IEKF update. These methods improve robustness while retaining the physical structure of model-based filters. However, they remain coupled to filter backends and, depending on the method, may also require explicit contact or slip estimates and manually defined thresholds. In contrast, purely learning-based odometry methods such as \textbf{Legolas}~\cite{wasserman2024legolas} directly predict the robot's odometry from IMU and leg histories without analytical kinematic models or filter-based physical constraints. The absence of such constraints can make the estimator more vulnerable to out-of-distribution motions and contact conditions, which can lead to large drift and estimation errors. Separately, because the learned mapping is optimized primarily on simulated sensor and motion distributions, discrepancies between simulation and the real robot can lead to a pronounced sim-to-real gap.

Motivated by these limitations, this paper proposes \textbf{TRACE}, an end-to-end learned estimator that provides proprioceptive odometry for legged robots operating under unreliable contact conditions. The estimator directly predicts relative displacement, relative rotation, and body-frame velocity from proprioceptive history, without relying on a filtering backend or manually defined contact and slip thresholds. To improve robustness to unreliable foot--ground interactions and reduce the sim-to-real gap, we jointly design the estimator architecture, training objectives, and two-stage training pipeline.

The main contributions of this paper are as follows:
\begin{itemize}
    \item We formulate quadruped proprioceptive odometry as an end-to-end learning problem that directly predicts relative motion and body-frame velocity from proprioceptive history.

    \item We develop a contact-aware estimator that couples foot-aware cross-attention with physics-inspired auxiliary objectives to downweight unreliable leg information without explicit contact or slip thresholds.

    \item We introduce a two-stage sim-to-real pipeline combining policy-randomized simulation pretraining with partial real-world fine-tuning using a closed-loop relative-error objective.

    \item We demonstrate consistent reductions in position drift over filtering-based, hybrid, and purely learning-based baselines across diverse indoor and outdoor terrains.
\end{itemize}

\section{Method}

\subsection{Estimator Input and Output}

The estimator receives a temporal window of proprioceptive observations
$\mathbf{o}_{t-T+1:t}$ with $T=30$. At each time step, the observation is defined as
\begin{equation}
    \begin{aligned}
    \mathbf{o}_t = \bigl\{&
    \boldsymbol{\omega}_t,\,\mathbf{a}_t,\,\mathbf{q}_t,\,\dot{\mathbf{q}}_t,\,\boldsymbol{\tau}_t,\,
    \hat{\theta}_{r,t-1},\,\hat{\theta}_{p,t-1},\,\\
    &\hat{v}_{x,t-1},\,\hat{v}_{y,t-1},\,\Delta t_t
    \bigr\}
    \in \mathbb{R}^{47}.
    \end{aligned}
    \label{eq:observation}
\end{equation}
Here, $\boldsymbol{\omega}_t \in \mathbb{R}^{3}$ and $\mathbf{a}_t \in \mathbb{R}^{3}$ denote the IMU angular velocity and linear acceleration measurements. The joint states are given by the joint positions $\mathbf{q}_t \in \mathbb{R}^{12}$, joint velocities $\dot{\mathbf{q}}_t \in \mathbb{R}^{12}$, and target joint torques $\boldsymbol{\tau}_t \in \mathbb{R}^{12}$. The observation also includes the previous roll and pitch estimates $\hat{\theta}_{r,t-1},\hat{\theta}_{p,t-1}$, previous horizontal body velocity estimates $\hat{v}_{x,t-1},\hat{v}_{y,t-1}$, and the sampling interval $\Delta t_t$. During training, these estimates are replaced with noisy ground-truth values obtained from simulation, whereas the estimator's previous outputs are used during fine-tuning and deployment.

Given this history, the network predicts a 9-dimensional output,
\begin{equation}
    \hat{\mathbf{y}}_t =
    \left[
    \Delta \hat{\mathbf{p}}_t^\top,
    \Delta \hat{\boldsymbol{\theta}}_t^\top,
    \hat{\mathbf{v}}_t^\top
    \right]^\top
    \in \mathbb{R}^{9},
    \label{eq:estimator_output}
\end{equation}
where $\Delta \hat{\mathbf{p}}_t \in \mathbb{R}^{3}$ is the body-frame relative displacement, $\Delta \hat{\boldsymbol{\theta}}_t \in \mathbb{R}^{3}$ is the relative rotation in the Lie algebra representation, and $\hat{\mathbf{v}}_t \in \mathbb{R}^{3}$ is the body-frame linear velocity.

The supervision label is computed from two consecutive robot states obtained from simulation. Let $\mathbf{p}_t \in \mathbb{R}^{3}$ be the body position in the world frame, $\mathbf{R}_t \in SO(3)$ be the body-to-world rotation matrix, and $\mathbf{v}^{W}_t \in \mathbb{R}^{3}$ be the world-frame body linear velocity. The relative displacement label is expressed in the previous body frame:
\begin{equation}
    \Delta \mathbf{p}_t
    =
    \mathbf{R}_{t-1}^{\top}
    \left(
    \mathbf{p}_t-\mathbf{p}_{t-1}
    \right).
    \label{eq:relative_displacement_label}
\end{equation}
The relative rotation from the previous body frame to the current body frame is represented in the Lie algebra as
\begin{equation}
    \Delta \boldsymbol{\theta}_t
    =
    \mathrm{Log}
    \left(
    \mathbf{R}_{t-1}^{\top}\mathbf{R}_t
    \right)^\vee .
    \label{eq:relative_rotation_label}
\end{equation}
The body-frame velocity label is obtained by rotating the world-frame velocity into the current body frame:
\begin{equation}
    \mathbf{v}_t
    =
    \mathbf{R}_t^{\top}\mathbf{v}^{W}_t .
    \label{eq:body_frame_velocity_label}
\end{equation}
Therefore, the training target is
\begin{equation}
    \mathbf{y}_t =
    \left[
    \Delta \mathbf{p}_t^\top,
    \Delta \boldsymbol{\theta}_t^\top,
    \mathbf{v}_t^\top
    \right]^\top
    \in \mathbb{R}^{9}.
    \label{eq:estimator_label}
\end{equation}

\subsection{Network Architecture}

\begin{figure}[t]
    \centering
    \includegraphics[width=\columnwidth]{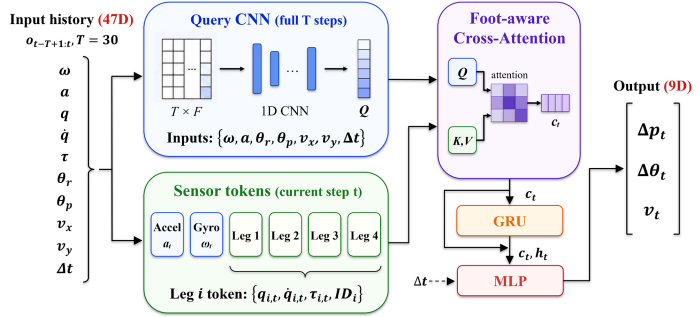}
    \caption{\textbf{Overview of the proposed foot-aware attention-based estimator.}
            A history-conditioned CNN query attends to current IMU and leg-wise tokens, and the resulting context is passed through a GRU and MLP head to predict relative displacement, relative rotation, and body-frame velocity.}
    \label{fig:architecture}
\end{figure}

The proposed estimator consists of four components: a proprioceptive query CNN, a foot-aware cross-attention module, a GRU temporal encoder, and an MLP prediction head. The overall architecture is shown in Fig.~\ref{fig:architecture}.

Inspired by convolutional sequence modeling~\cite{bai2018empirical}, a temporal CNN encodes recent body-motion signals into a compact query feature $\mathbf{Q}_t$. The CNN input consists of a $T=30$-sample window of $\{\boldsymbol{\omega}, \mathbf{a}, \hat{\theta}_r, \hat{\theta}_p, \hat{v}_x, \hat{v}_y, \Delta t\}$. The query encoder uses two 1D convolutional layers, followed by temporal pooling, a linear projection, and layer normalization. The resulting $d$-dimensional feature summarizes recent body motion and serves as a history-conditioned query for the foot-aware attention module.

The cross-attention module attends to six sensor tokens constructed from the current proprioceptive measurements: linearly embedded acceleration and gyroscope tokens and four leg tokens. Each leg token embeds the 13-dimensional feature $\{q_i, \dot{q}_i, \tau_i, ID_i\}$, consisting of joint position, joint velocity, target joint torque, and a one-hot leg ID, using a small MLP. The CNN query and the six sensor tokens are then processed by a two-head cross-attention module with embedding dimension $d$. For attention head $h$, the weight assigned to the $i$-th token and the corresponding context vector are computed as
\begin{equation}
    \alpha^{(h)}_{i,t}
    =
    \mathrm{softmax}_i
    \left(
    \frac{
    {\mathbf{Q}^{(h)}_t}^{\top}\mathbf{k}^{(h)}_{i,t}
    }{\sqrt{d_h}}
    \right),
    \mathbf{c}^{(h)}_t
    =
    \sum_i \alpha^{(h)}_{i,t}\mathbf{v}^{(h)}_{i,t},
    \label{eq:cross_attention}
\end{equation}
where $d_h$ is the head dimension. The head-wise context vectors are combined to form $\mathbf{c}_t$, and the token attention used in the foot velocity loss and analysis is $\alpha_{i,t}=\frac{1}{H}\sum_{h=1}^{H}\alpha^{(h)}_{i,t}$, with $H=2$. This structure allows the network to adaptively emphasize different IMU and leg features depending on the current locomotion condition.

The attention context is then passed to a GRU with a hidden dimension of 128. An MLP with hidden dimensions [256, 128] uses the attention context, GRU hidden state, and sampling interval to predict the 9-dimensional output. The shared embedding dimension is $d=64$. During evaluation and deployment, a ZUPT-inspired output clamp sets the predicted motion to zero when the angular and joint velocities are small and the measured acceleration is close to gravity, thereby suppressing stationary drift.

\subsection{Training Losses}

The total training objective consists of estimation loss, kinematic model consistency loss, and foot velocity loss:
\begin{equation}
    \mathcal{L}
    =
    w_{\mathrm{est}} \mathcal{L}_{\mathrm{est}}
    +
    w_{\mathrm{model}} \mathcal{L}_{\mathrm{model}}
    +
    w_{\mathrm{foot}} \mathcal{L}_{\mathrm{foot}}.
    \label{eq:total_loss}
\end{equation}

\subsubsection{Estimation Loss}

The estimation loss directly supervises the predicted relative displacement, relative rotation, and velocity using Smooth L1 loss:
\begin{equation}
    \mathcal{L}_{\mathrm{est}}
    =
    w_p \rho(\Delta \hat{\mathbf{p}}_t - \Delta \mathbf{p}_t)
    +
    w_R \rho(\Delta \hat{\boldsymbol{\theta}}_t - \Delta \boldsymbol{\theta}_t)
    +
    w_v \rho(\hat{\mathbf{v}}_t - \mathbf{v}_t),
    \label{eq:estimation_loss}
\end{equation}
where $\rho(\cdot)$ denotes the Smooth L1 loss.

\subsubsection{Kinematic Model Consistency Loss}

The model consistency loss encourages consistency between the predicted displacement and the displacement reconstructed from predicted velocity. Using trapezoidal integration, the velocity-based displacement is approximated as
\begin{equation}
    \Delta \mathbf{p}^{\mathrm{kin}}_{t}
    =
    \frac{\Delta t_{t}}{2}
    \left(
    \mathbf{v}^{\mathrm{gt}}_{t-1}
    +
    \Delta\hat{\mathbf{R}}_{t}\hat{\mathbf{v}}_{t}
    \right).
    \label{eq:kinematic_displacement}
\end{equation}
The model consistency loss is then defined as
\begin{equation}
    \mathcal{L}_{\mathrm{model}}
    =
    \rho(
    \Delta \hat{\mathbf{p}}_t
    -
    \Delta \mathbf{p}^{\mathrm{kin}}_{t}
    ).
    \label{eq:model_consistency_loss}
\end{equation}

This loss enforces kinematic consistency among the predicted rotation, velocity, and displacement.

\subsubsection{Foot Velocity Loss}

\begin{figure}[t]
    \centering
    \includegraphics[width=\columnwidth]{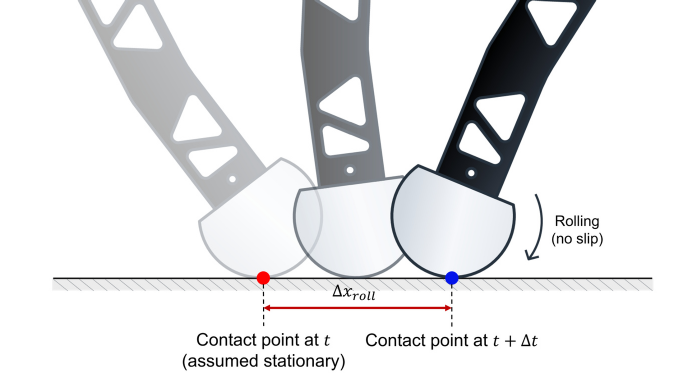}
    \caption{\textbf{Contact-location shift under no-slip rolling.}
    The overlaid configurations illustrate how rolling shifts the instantaneous contact location from the initial point (red) to the later point (blue), violating the fixed-foot-point assumption.}
    \label{fig:foot_rolling}
\end{figure}

The foot velocity loss promotes physically consistent contact behavior. Contact-aided filters often assume that a kinematically defined foot point remains stationary during stance~\cite{bloesch2013state,hartley2020contact,lee2026attention}. However, as illustrated in Fig.~\ref{fig:foot_rolling}, foot rolling shifts the instantaneous contact location even in the absence of slip and can lead to displacement underestimation. We therefore compute the contact-point velocity of each leg using foot kinematics and contact normals available in simulation and weight its penalty by the corresponding leg attention.

For each leg, forward kinematics provides the foot-center position $\mathbf{p}^{\mathrm{foot}}_{i,t}$. Let $\mathbf{J}^{v}_{i}(\mathbf{q}_{i,t})$ and $\mathbf{J}^{\omega}_{i}(\mathbf{q}_{i,t})$ denote the linear and angular parts of the foot-link geometric Jacobian expressed in the body frame, respectively. The foot-center linear velocity is
\begin{equation}
    \mathbf{v}^{\mathrm{foot}}_{i,t}
    =
    \hat{\mathbf{v}}_{t}
    +
    \boldsymbol{\omega}_{t}
    \times
    \mathbf{p}^{\mathrm{foot}}_{i,t}
    +
    \mathbf{J}^{v}_{i}(\mathbf{q}_{i,t})
    \dot{\mathbf{q}}_{i,t}.
    \label{eq:foot_center_velocity}
\end{equation}
The angular velocity of the foot link is
\begin{equation}
    \boldsymbol{\omega}^{\mathrm{foot}}_{i,t}
    =
    \boldsymbol{\omega}_{t}
    +
    \mathbf{J}^{\omega}_{i}(\mathbf{q}_{i,t})
    \dot{\mathbf{q}}_{i,t}.
\end{equation}
For a valid ground normal $\mathbf{n}_{i,t}$ pointing toward the foot, the offset from the foot center to the contact point is $\mathbf{r}^{\mathrm{contact}}_{i,t} =-r_{\mathrm{foot}}\mathbf{n}_{i,t}/\|\mathbf{n}_{i,t}\|_2$. The contact-point velocity is then \begin{equation} 
    \mathbf{v}^{\mathrm{contact}}_{i,t} 
    = 
    \mathbf{v}^{\mathrm{foot}}_{i,t} 
    + 
    \boldsymbol{\omega}^{\mathrm{foot}}_{i,t} 
    \times 
    \mathbf{r}^{\mathrm{contact}}_{i,t}. 
    \label{eq:contact_velocity_compact}
\end{equation}
If there is no valid contact normal, we use $\mathbf{v}^{\mathrm{contact}}_{i,t}=\mathbf{v}^{\mathrm{foot}}_{i,t}$.
Let $A_t=\sum_{j=1}^{4}\alpha_{j,t}^{leg}$ and $\bar{\alpha}^{leg}_{i,t}=\alpha_{i,t}^{leg}/(A_t+\epsilon)$. The foot velocity loss is
\begin{equation}
    \mathcal{L}_{\mathrm{foot}}
    =
    \mathrm{sg}(A_t)
    \sum_{i=1}^{4}
    \bar{\alpha}_{i,t}^{leg}
    \left\|
    \mathbf{v}^{\mathrm{contact}}_{i,t}
    \right\|_2 ,
    \label{eq:foot_velocity_loss_compact}
\end{equation}
where $\mathrm{sg}(\cdot)$ denotes stop-gradient. The objective penalizes large contact-point velocities in proportion to leg attention, encouraging reliance on kinematically reliable legs. The stop-gradient on $A_t$ prevents trivial loss reduction through collapse of the total leg attention.

\subsection{Simulation Training Details}

The estimator is trained in RaiSim~\cite{hwangbo2018per} using trajectories generated by a pretrained policy. We simulate 400 parallel environments with 4~s rollouts, recording estimator inputs, labels and contact normals. The policy, estimator, and simulator operate at 100~Hz, 500~Hz, and 4~kHz, respectively. Training uses truncated backpropagation through time with a sequence length of 100 and Adam~\cite{kingma2014adam} with a learning rate of $3\times10^{-4}$. We perform 3000 updates with 10 epochs per rollout batch, requiring approximately 20 hours on an NVIDIA RTX 4060 GPU.

At the beginning of each rollout, the robot is commanded to stand with approximately 10\% probability; otherwise, a random locomotion command is applied. Additionally, we apply domain randomization over terrain, contact, dynamics, sensors, timing, and external disturbances. Terrain randomization includes flat ground, Perlin and discrete height maps, random and inclined steps, and stairs following Lee et al.~\cite{lee2020learning}. Ground friction is sampled as $\mu\sim\mathcal{U}(0.4,1.2)$, while sudden slip is simulated by reducing the friction of a contacted foot to $\mu_{\mathrm{slip}}\sim\mathcal{U}(0.3,0.4)$ with probability $1\%$ until contact is lost. Dynamics randomization perturbs inertial properties, joint friction, and motor saturation, while sensor and timing randomization vary sensor measurements and sampling intervals. We additionally apply external torso disturbances to expose the estimator to transient motions outside the nominal locomotion distribution.

\subsection{Policy Randomization}

A proprioceptive estimator trained using a single fixed locomotion policy may overfit to the gait timing, action distribution, and contact patterns induced by that policy. This issue is particularly important for end-to-end learned odometry, because sim-to-real discrepancies can cause the same locomotion policy to produce different motion patterns on the real robot, introducing an additional distribution shift that exacerbates estimation errors. To address this issue, we introduce policy randomization (PR), which broadens the range of locomotion behaviors encountered during simulation data collection while retaining a single pretrained policy. Specifically, the action scale is sampled as \(0.1 \times \mathcal{U}(0.7, 1.2)\,\mathrm{rad}\), the action mean is perturbed by a Gaussian bias sampled from
\(\mathcal{N}\!\left(0, (0.006\,\mathrm{rad})^2\right)\), and the target actions are low-pass filtered using a coefficient \(\alpha \sim \mathcal{U}(0, 0.6)\). Policy randomization diversifies foot-contact timing, step length, impact profiles, and body-motion distributions even under similar velocity commands. Consequently, the estimator is encouraged to learn robust proprioceptive patterns rather than policy-specific motion artifacts, reducing its dependence on the training policy and improving sim-to-real generalization.

\subsection{Real-World Fine-Tuning}

After simulation pretraining, the estimator is fine-tuned on real-world outdoor trajectories to reduce the sim-to-real gap. We use six FAST-LIO2~\cite{xu2022fast}-based $150~\mathrm{s}$ training logs, three collected on grass and three on flat ground, under random velocity commands with maximum speeds of $1$--$3~\mathrm{m/s}$. To augment the training data, each log is divided into overlapping $20~\mathrm{s}$ windows with a $10~\mathrm{s}$ stride, and the estimator is rolled out in closed loop within each window. To avoid overfitting, only the state GRU and MLP head are updated, while the CNN query encoder and attention module are frozen. Fine-tuning uses Adam with a learning rate of $1\times10^{-5}$ for 20 epochs, with gradient norm clipping at 1.0.

Real-world fine-tuning uses a $1~\mathrm{s}$-window relative error objective instead of per-step supervision. For each frame, an anchor frame $1~\mathrm{s}$ earlier is selected, and the predicted and reference relative motions are compared in the anchor body frame. To decouple the RE position term from rotation prediction, the predicted translation is rolled out using FAST-LIO2 reference rotations. The relative error loss is defined as
$\mathcal{L}_{\mathrm{RE}} = w_{\mathrm{pos}}\mathcal{L}^{\mathrm{RE}}_{\mathrm{pos}} + w_{\mathrm{rot}}\mathcal{L}^{\mathrm{RE}}_{\mathrm{rot}}$, and the final objective is

\begin{equation}
    \mathcal{L}_{\mathrm{FT}}
    =
    w_{\mathrm{RE}}\mathcal{L}_{\mathrm{RE}}
    +
    w_{\mathrm{model}}\mathcal{L}_{\mathrm{model}}
    +
    w_{\mathrm{SP}}\mathcal{L}_{\mathrm{SP}} .
    \label{eq:finetune_loss}
\end{equation}

Here, $\mathcal{L}_{\mathrm{SP}}$ is the $L^{2}$-SP regularization term that penalizes deviation from the simulation-pretrained weights~\cite{xuhong2018explicit}. To also improve velocity estimation, we retain the model consistency loss during fine-tuning, replacing $\mathbf{v}^{\mathrm{gt}}_{t-1}$ in~\eqref{eq:kinematic_displacement} with the previously predicted velocity. Direct velocity supervision is excluded because real-world velocity labels are noisy and empirically did not improve performance.

\section{Experimental Results}

\subsection{Experimental Setup}

For real-world evaluation, we use a quadruped robot Raibo2~\cite{hwangbo2025raibo2}, equipped with an onboard IMU and joint encoders. The IMU measurements and joint states are recorded at 500~Hz and the proposed estimator as well as the baselines are evaluated offline using the recorded sensor data. Although the evaluation is performed offline for fair comparison under identical input sequences, the proposed estimator is sufficiently lightweight for online deployment. We measured its inference latency on an Intel Core Ultra 7 255H CPU over 10,000 sequential samples with batch size one, obtaining an average of $0.2774 \pm 0.0188~\mathrm{ms}$ per step, which is sufficient for 500~Hz operation.

\begin{figure}[t]
    \centering
    \includegraphics[width=\columnwidth]{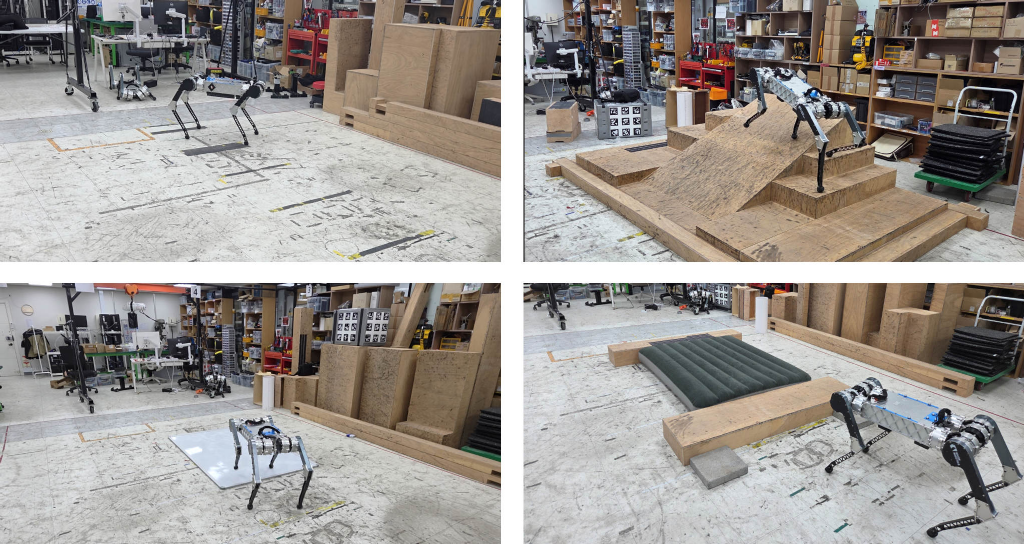}
    \caption{\textbf{Indoor Experiment Setup.} Upper left - \textit{Flat}, Upper right - \textit{Rough}, Lower left - \textit{Slippery}, Lower right - \textit{Soft}.}
    \label{fig:terrains}
\end{figure}

Indoor evaluation is conducted on \textit{flat, rough, slippery,} and \textit{soft} terrains, as shown in Fig.~\ref{fig:terrains}. Vicon ground-truth trajectories are recorded at 200~Hz using 10 Vero V2.2 cameras and interpolated to 500~Hz. Outdoor evaluation includes \textit{grass, hard ground, stairs,} and a \textit{full-course trajectory}, with FAST-LIO2 reference trajectories recorded at 10~Hz and interpolated to 500~Hz. Since FAST-LIO2 is evaluated primarily in terms of translational accuracy in~\cite{xu2022fast}, only position ATE and RE are reported. All evaluation trajectories are disjoint from the fine-tuning data.

\subsection{Baselines and Metrics}

We compare the proposed estimator with the following three baselines:
\begin{itemize}
    \item \textbf{IEKF-SR}: a classic GRF contact-aided IEKF with slip rejection following a method proposed in~\cite{kim2021legged}.

    \item \textbf{Legolas}: a purely learning-based odometry estimator that predicts pose increments, but not velocity. For a fair comparison, it is trained in the same simulation environment as ours and adapted to operate at 500~Hz~\cite{wasserman2024legolas}.

    \item \textbf{NMN-IEKF}: a hybrid estimator that uses network-predicted contact probabilities and body-frame linear velocity within an IEKF, trained following the domain-randomization protocol of the original work~\cite{youm2025legged}.
\end{itemize}

We report absolute trajectory error (ATE) and 10-second relative error (RE) for position, velocity, and orientation, following the trajectory evaluation protocol in~\cite{zhang2018tutorial}. Position, velocity, and orientation errors are reported in meters (m), meters per second (m/s), and radians (rad), respectively. For the outdoor experiments, we additionally report the standard deviation and 90th percentile of the 10-second position RE to better characterize estimation robustness and tail-error behavior.

\subsection{Attention Analysis}

For the attention analysis, we use simulation rollouts to verify whether the proposed foot-aware attention module learns meaningful foot reliability. The analysis uses 2,000,000 per-foot samples, consisting of 1,129,612 contact samples and 870,388 swing samples.

\begin{table}[t]
    \centering
    \caption{\textbf{Attention Analysis}: Learned attention aligns with foot contact state}
    \label{tab:attention_contact_analysis}
    \renewcommand{\arraystretch}{1.15}
    \setlength{\tabcolsep}{5pt}
    \begin{tabular}{|c|c|c|}
        \hline
        \textbf{Metric} & \textbf{Contact} & \textbf{Swing} \\
        \hline
        Mean per-foot attention & 0.2106 & 0.0060 \\
        \hline
        Median per-foot attention & 0.1984 & 0.0000 \\
        \hline
        Top-1 attention on contact foot & \multicolumn{2}{c|}{96.21\%} \\
        \hline
        AUC(contact label, attention) & \multicolumn{2}{c|}{0.9637} \\
        \hline
    \end{tabular}
\end{table}

\begin{figure}[t]
    \centering
    \includegraphics[width=\columnwidth]{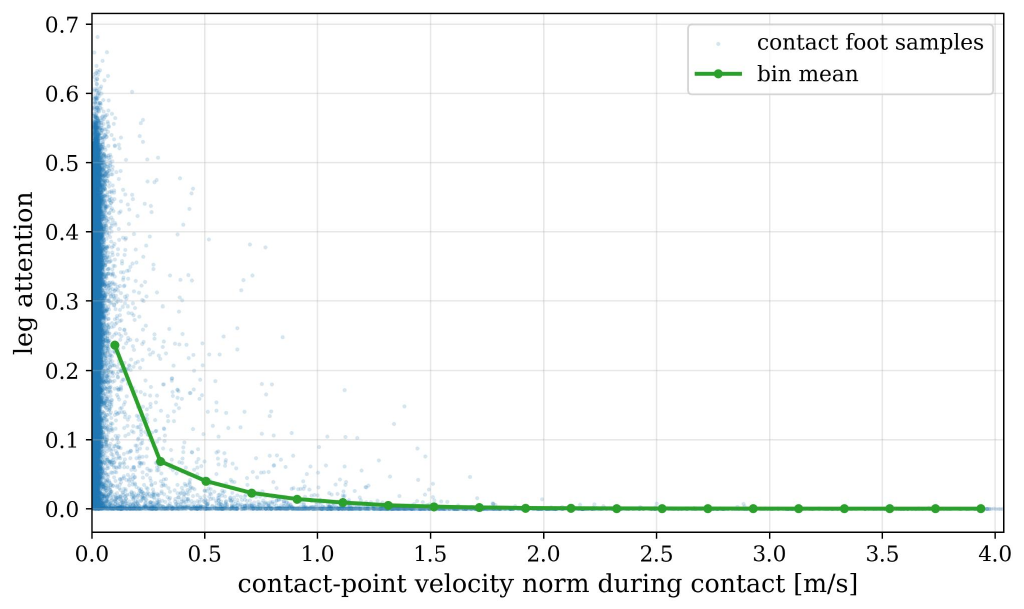}
    \caption{\textbf{Relationship between leg attention and contact-point velocity during contact.} The binned mean attention decreases as contact-point velocity increases, indicating that the network assigns lower attention to unreliable contact feet.}
    \label{fig:attention_analysis}
\end{figure}

As shown in Table~\ref{tab:attention_contact_analysis}, the learned attention strongly aligns with contact state: the highest-attention foot is in contact in 96.21\% of samples, and contact feet receive substantially higher mean and median attention than swing feet. The attention score also distinguishes contact from swing with an AUC of 0.9637, showing that foot-velocity supervision induces the attention module to emphasize physically informative stance legs without using any explicit contact threshold. Figure~\ref{fig:attention_analysis} further shows that attention decreases rapidly as contact-point velocity increases. This trend suggests that the foot velocity loss shapes attention into a soft reliability measure that downweights contact information contaminated by slip, deformation, or transient contact dynamics.

\subsection{Indoor Terrain Experiments}

We evaluate the proposed estimator on four indoor terrain conditions.

\begin{itemize}
    \item \textbf{\textit{Flat}}: The robot traverses a rectangular trajectory of 39.37 meters on normal flat ground for 60 seconds.

    \item \textbf{\textit{Rough}}: We construct a rough terrain using wooden platforms. The terrain consists of inclined ramps, elevated blocks, and stair-like structures, inducing large pitch variations and intermittent foot contact during locomotion. The robot traverses the terrain for 60 seconds.

    \item \textbf{\textit{Slippery}}: We spread boric acid powder on a white board to create a low-friction surface, reducing the friction coefficient to below 0.3. The robot traverses a round-trip trajectory of 32.30 meters for 60 seconds.

    \item \textbf{\textit{Soft}}: We place an air mat on the ground to create a soft terrain which is not modeled during simulation training. The robot traverses a 29.21 m round-trip trajectory for 60 seconds.
\end{itemize}

\begin{table}[t]
    \centering
    \caption{ATE and 10-second RE Across Various Indoor Terrain Experiments}
    \label{tab:indoor_results}
    \renewcommand{\arraystretch}{1.15}
    \setlength{\tabcolsep}{3pt}
    \resizebox{\columnwidth}{!}{
    \begin{tabular}{|c|c|c|c|c|c|c|c|}
        \hline
        \multirow{2}{*}{\textbf{Terrain}}
        & \multirow{2}{*}{\textbf{Method}}
        & \textbf{ATE} & \textbf{ATE} & \textbf{ATE}
        & \textbf{RE} & \textbf{RE} & \textbf{RE} \\
        &
        & \textbf{(pos)} & \textbf{(vel)} & \textbf{(ori)}
        & \textbf{(pos)} & \textbf{(vel)} & \textbf{(ori)} \\
        \hline

        \multirow{4}{*}{\textit{Flat}}
        & \textbf{Proposed}
        & \textbf{0.0882} & \textbf{0.0561} & \textbf{0.0352}
        & \textbf{0.0538} & 0.0795 & \textbf{0.0172} \\
        \cline{2-8}
        & IEKF-SR
        & 0.3673 & 0.0692 & 0.1758
        & 0.1742 & 0.0992 & 0.0569 \\
        \cline{2-8}
        & Legolas
        & 0.3886 & \textemdash & 0.1272
        & 0.1357 & \textemdash & 0.0534 \\
        \cline{2-8}
        & NMN-IEKF
        & 0.1788 & 0.0582 & 0.0511
        & 0.0769 & \textbf{0.0789} & 0.0205 \\
        \hline

        \multirow{4}{*}{\textit{Rough}}
        & \textbf{Proposed}
        & \textbf{0.1656} & \textbf{0.1018} & \textbf{0.0221}
        & \textbf{0.0977} & \textbf{0.1224} & \textbf{0.0199} \\
        \cline{2-8}
        & IEKF-SR
        & 0.2722 & 0.1053 & 0.1294
        & 0.2028 & 0.1395 & 0.0425 \\
        \cline{2-8}
        & Legolas
        & 0.8531 & \textemdash & 0.2482
        & 0.3050 & \textemdash & 0.0894 \\
        \cline{2-8}
        & NMN-IEKF
        & 0.5755 & 0.1265 & 0.0422
        & 0.1988 & 0.1361 & 0.0214 \\
        \hline

        \multirow{4}{*}{\textit{Slippery}}
        & \textbf{Proposed}
        & \textbf{0.1154} & \textbf{0.0672} & \textbf{0.0246}
        & \textbf{0.0550} & \textbf{0.0758} & \textbf{0.0128} \\
        \cline{2-8}
        & IEKF-SR
        & 0.2980 & 0.0793 & 0.1548
        & 0.1845 & 0.1109 & 0.0471 \\
        \cline{2-8}
        & Legolas
        & 0.6849 & \textemdash & 0.3473
        & 0.2715 & \textemdash & 0.1045 \\
        \cline{2-8}
        & NMN-IEKF
        & 0.2065 & 0.0691 & 0.0423
        & 0.0915 & 0.0790 & 0.0173 \\
        \hline

        \multirow{4}{*}{\textit{Soft}}
        & \textbf{Proposed}
        & \textbf{0.1530} & 0.0844 & \textbf{0.0153}
        & \textbf{0.0930} & \textbf{0.0941} & \textbf{0.0133} \\
        \cline{2-8}
        & IEKF-SR
        & 0.7027 & \textbf{0.0801} & 0.1083
        & 0.2878 & 0.1017 & 0.0380 \\
        \cline{2-8}
        & Legolas
        & 0.7620 & \textemdash & 0.2596
        & 0.3304 & \textemdash & 0.0949 \\
        \cline{2-8}
        & NMN-IEKF
        & 0.3309 & 0.0919 & 0.0241
        & 0.1264 & 0.1001 & 0.0143 \\
        \hline
    \end{tabular}
    }
\end{table}

\begin{figure*}[t]
    \centering
    \includegraphics[width=\textwidth]{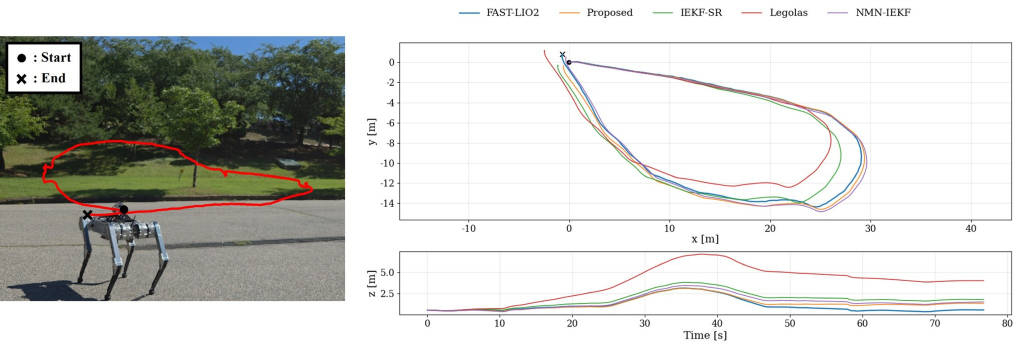}
    \caption{\textbf{Long-horizon outdoor trajectory comparison.}
    Left: the traveled path overlaid on an image of the experimental environment. Right: the top-down $xy$ trajectories (top) and vertical position profiles over time (bottom). The proposed estimator closely reconstructs both the horizontal trajectory and the reference elevation profile, demonstrating consistent accuracy across diverse real-world environments.}
    \label{fig:trajectory_comparison}
\end{figure*}

Table~\ref{tab:indoor_results} summarizes the indoor evaluation results. The proposed estimator achieves the lowest position and orientation errors across all terrain conditions, demonstrating consistent robustness under both nominal and unreliable contact scenarios. Compared with the strongest baseline for each terrain, the proposed method
reduces the position ATE by \textbf{50.7\%, 39.2\%, 44.1\%, and 53.8\%} on flat, rough, slippery, and soft terrains, respectively. A similar trend is observed for the 10-second position RE, with reductions of \textbf{30.0\%, 50.9\%, 39.9\%, and 26.4\%}, respectively. These gains are especially meaningful on soft terrain, where the unmodeled foot sinkage and terrain deformation violate rigid-contact assumptions.

Compared with the model-based baselines, the proposed estimator shows a larger improvement in position estimation than in velocity estimation. This discrepancy is expected because filter-based estimators tightly couple velocity and position increments through their propagation and update equations, whereas the proposed end-to-end estimator predicts displacement and velocity jointly and only encourages their consistency through the model consistency loss. Therefore, improvements in accumulated position drift and instantaneous velocity estimation do not necessarily coincide. Nevertheless, the proposed estimator achieves lower velocity error on most terrains.

\subsection{Outdoor Terrain Experiments}

We evaluate the proposed estimator on four outdoor conditions.

\begin{itemize}
    \item \textbf{\textit{Grass}}: The robot traverses a rectangular trajectory of 57.63 meters on grass for 60 seconds. This terrain introduces friction variation and contact uncertainty due to the compliant and uneven surface.

    \item \textbf{\textit{Hard ground}}: The robot traverses a rectangular trajectory of 58.69 meters on hard outdoor ground for 60 seconds.

    \item \textbf{\textit{Stairs}}: The robot climbs and descends stairs for 60 seconds.

    \item \textbf{\textit{Full course}}: The robot traverses a full outdoor course of 72.22 meters for 76 seconds. The course consists of inclined grass terrain, hard ground, and several steps.
\end{itemize}

\begin{table}[t]
    \centering
    \caption{Position ATE and 10-second Position RE Across Outdoor Experiments}
    \label{tab:outdoor_results}
    \renewcommand{\arraystretch}{1.15}
    \setlength{\tabcolsep}{2pt}
    \footnotesize
    \renewcommand{\tabularxcolumn}[1]{m{#1}}

    \begin{tabularx}{\columnwidth}{
        |>{\centering\arraybackslash}m{0.13\columnwidth}
        |>{\centering\arraybackslash}m{0.19\columnwidth}
        |>{\hsize=0.75\hsize\centering\arraybackslash}X
        |>{\hsize=1.45\hsize\centering\arraybackslash}X
        |>{\hsize=0.80\hsize\centering\arraybackslash}X|
    }
        \hline
        \textbf{Terrain}
        & \textbf{Method}
        & \textbf{ATE (pos)}
        & \shortstack[c]{\rule{0pt}{2.5ex}\textbf{RE (pos)}\\mean $\pm$ std}
        & \shortstack[c]{\rule{0pt}{2.5ex}\textbf{RE (pos)}\\P90} \\
        \hline

        \multirow{4}{*}{\textit{Grass}}
        & \textbf{Proposed}
        & \textbf{0.4831}
        & $\boldsymbol{0.1633 \pm 0.0896}$
        & \textbf{0.2620} \\
        \cline{2-5}
        & IEKF-SR
        & 1.0078
        & $0.5592 \pm 0.0893$
        & 0.6538 \\
        \cline{2-5}
        & Legolas
        & 2.3665
        & $0.7466 \pm 0.1222$
        & 0.9028 \\
        \cline{2-5}
        & NMN-IEKF
        & 0.5820
        & $0.3395 \pm 0.0722$
        & 0.4299 \\
        \hline

        \multirow{4}{*}{\textit{\shortstack{Hard\\ground}}}
        & \textbf{Proposed}
        & \textbf{0.1796}
        & $\boldsymbol{0.1738 \pm 0.1617}$
        & \textbf{0.4144} \\
        \cline{2-5}
        & IEKF-SR
        & 0.9290
        & $0.5453 \pm 0.1600$
        & 0.7389 \\
        \cline{2-5}
        & Legolas
        & 1.4581
        & $0.5443 \pm 0.2409$
        & 0.8001 \\
        \cline{2-5}
        & NMN-IEKF
        & 0.3166
        & $0.2597 \pm 0.1550$
        & 0.5230 \\
        \hline

        \multirow{4}{*}{\textit{Stairs}}
        & \textbf{Proposed}
        & \textbf{0.1629}
        & $\boldsymbol{0.1458 \pm 0.0695}$
        & \textbf{0.2194} \\
        \cline{2-5}
        & IEKF-SR
        & 0.4421
        & $0.2987 \pm 0.1276$
        & 0.4758 \\
        \cline{2-5}
        & Legolas
        & 1.3512
        & $0.4247 \pm 0.2413$
        & 0.7691 \\
        \cline{2-5}
        & NMN-IEKF
        & 0.4445
        & $0.1506 \pm 0.0613$
        & 0.2292 \\
        \hline

        \multirow{4}{*}{\textit{\shortstack{Full\\course}}}
        & \textbf{Proposed}
        & \textbf{0.7328}
        & $\boldsymbol{0.2078 \pm 0.1396}$
        & \textbf{0.3637} \\
        \cline{2-5}
        & IEKF-SR
        & 1.5379
        & $0.6194 \pm 0.1011$
        & 0.6998 \\
        \cline{2-5}
        & Legolas
        & 3.8345
        & $0.8831 \pm 0.1579$
        & 1.0639 \\
        \cline{2-5}
        & NMN-IEKF
        & 0.7898
        & $0.3321 \pm 0.1527$
        & 0.5249 \\
        \hline
    \end{tabularx}
\end{table}

Fig.~\ref{fig:trajectory_comparison} shows that the proposed estimator accurately reconstructs the full trajectory over the long outdoor sequence. On the full course, it achieves the lowest mean 10-s position RE and P90, reducing them by \textbf{37.4\% and 30.7\%} relative to the strongest baseline, respectively, while maintaining competitive position ATE. This result demonstrates that the estimator generalizes reliably across diverse outdoor terrains and maintains robust performance under uncontrolled real-world conditions.

The terrain-wise results in Table~\ref{tab:outdoor_results} further support this observation. Compared with the strongest baseline for each metric, the proposed method reduces position ATE by \textbf{17.0\%, 43.3\%, and 63.2\%} on grass, hard ground, and stairs, respectively, and reduces mean 10-s position RE by \textbf{51.9\%, 33.1\%, and 3.2\%}. The substantial RE improvement on grass demonstrates robustness to compliant and deformable contacts, while the large ATE reduction on stairs indicates improved consistency under repeated impacts and rapid elevation changes. The lower P90 values further indicate improved robustness to large estimation errors under uneven terrain and compliant contacts.

\subsection{Ablation Study}

We conduct ablation studies to answer the following questions. First, do the proposed auxiliary objectives improve estimation accuracy and help identify reliable leg information? Second, does policy randomization reduce policy-specific overfitting and improve sim-to-real robustness? Third, does the proposed real-world fine-tuning strategy reduce the sim-to-real gap while preserving generalization?

\begin{table}[t]
    \centering
    \caption{\textbf{Ablation Study I}: Impact of Auxiliary Losses}
    \label{tab:auxiliary_loss_ablation}
    \renewcommand{\arraystretch}{1.15}
    \setlength{\tabcolsep}{3pt}
    \resizebox{\columnwidth}{!}{
    \begin{tabular}{|c|c|c|c|c|c|c|c|}
        \hline
        \multirow{2}{*}{\textbf{Terrain}}
        & \multirow{2}{*}{\textbf{Method}}
        & \textbf{ATE} & \textbf{ATE} & \textbf{ATE}
        & \textbf{RE} & \textbf{RE} & \textbf{RE} \\
        & & \textbf{(pos)} & \textbf{(vel)} & \textbf{(ori)}
        & \textbf{(pos)} & \textbf{(vel)} & \textbf{(ori)} \\
        \hline

        \multirow{3}{*}{\textit{Flat}}
        & \textbf{Proposed}
        & \textbf{0.0882} & \textbf{0.0561} & 0.0352
        & \textbf{0.0538} & \textbf{0.0795} & 0.0172 \\
        \cline{2-8}
        & w/o $\mathcal{L}_{\mathrm{model}}$
        & 0.1926 & 0.0809 & 0.0465
        & 0.0685 & 0.0971 & 0.0181 \\
        \cline{2-8}
        & w/o both
        & 0.7422 & 0.1179 & \textbf{0.0123}
        & 0.5826 & 0.2110 & \textbf{0.0138} \\
        \hline

        \multirow{3}{*}{\textit{Rough}}
        & \textbf{Proposed}
        & \textbf{0.1656} & \textbf{0.1018} & \textbf{0.0221}
        & \textbf{0.0977} & \textbf{0.1224} & \textbf{0.0199} \\
        \cline{2-8}
        & w/o $\mathcal{L}_{\mathrm{model}}$
        & 0.1820 & 0.1482 & 0.0332
        & 0.1092 & 0.1580 & 0.0219 \\
        \cline{2-8}
        & w/o both
        & 1.6707 & 0.1785 & 0.0317
        & 0.7678 & 0.2442 & 0.0212 \\
        \hline

        \multirow{3}{*}{\textit{Slippery}}
        & \textbf{Proposed}
        & \textbf{0.1154} & \textbf{0.0672} & \textbf{0.0246}
        & \textbf{0.0550} & \textbf{0.0758} & \textbf{0.0128} \\
        \cline{2-8}
        & w/o $\mathcal{L}_{\mathrm{model}}$
        & 0.1265 & 0.0935 & 0.0391
        & 0.0771 & 0.1235 & 0.0171 \\
        \cline{2-8}
        & w/o both
        & 0.6484 & 0.1171 & 0.0464
        & 0.3624 & 0.1532 & 0.0187 \\
        \hline

        \multirow{3}{*}{\textit{Soft}}
        & \textbf{Proposed}
        & \textbf{0.1530} & \textbf{0.0844} & \textbf{0.0153}
        & \textbf{0.0930} & \textbf{0.0941} & \textbf{0.0133} \\
        \cline{2-8}
        & w/o $\mathcal{L}_{\mathrm{model}}$
        & 0.2280 & 0.1123 & 0.0256
        & 0.1095 & 0.1207 & 0.0204 \\
        \cline{2-8}
        & w/o both
        & 1.4326 & 0.2125 & 0.0252
        & 1.0845 & 0.3610 & 0.0191 \\
        \hline
    \end{tabular}
    }
\end{table}

\begin{figure}[t]
    \centering
    \includegraphics[width=\columnwidth]{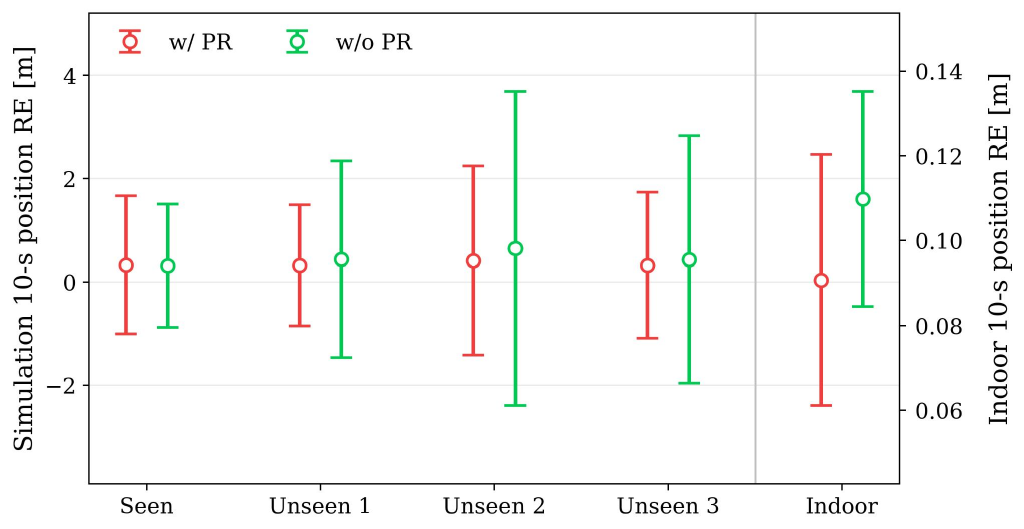}
    \caption{\textbf{Effect of policy randomization on policy and sim-to-real generalization.}
    The left axis shows the 10-s position RE in simulation for the policy used during estimator training and three independently trained unseen policies, while the right axis shows the zero-shot indoor result averaged over four real-world terrains. Markers indicate the mean, and error bars denote one standard deviation.}
    \label{fig:PR_ablation}
\end{figure}

Table~\ref{tab:auxiliary_loss_ablation} evaluates the contributions of the auxiliary objectives. The w/o $\mathcal{L}_{\mathrm{model}}$ variant removes $\mathcal{L}_{\mathrm{model}}$ from both simulation pretraining and real-world fine-tuning, while w/o both additionally removes $\mathcal{L}_{\mathrm{foot}}$ during simulation. The complete model reduces the mean position and velocity ATE by 28.4\% and 28.8\%, respectively, compared with w/o $\mathcal{L}_{\mathrm{model}}$, confirming the benefit of kinematic consistency. Adding $\mathcal{L}_{\mathrm{foot}}$ to w/o both reduces the mean position ATE and RE by 83.8\% and 87.0\%, respectively, and the velocity ATE and RE by 30.5\% and 48.5\%. This improvement is accompanied by a reduction in mean swing-foot attention from 0.0452 to 0.0060, indicating that the loss selectively suppresses unreliable leg information and thereby reduces velocity bias and accumulated position drift. These results demonstrate that the two auxiliary objectives provide complementary benefits for physically consistent and robust proprioceptive odometry.

To evaluate policy randomization, we compare the proposed model with w/o PR on the training policy and three independently trained unseen policies with the same gait. As shown in Fig.~\ref{fig:PR_ablation}, w/o PR performs marginally better on the seen policy, whereas the proposed model achieves lower mean error and lower standard deviation across all unseen policies. Averaged over the three unseen policies, policy randomization reduces the 10-s position RE from 0.5076 to 0.3516~m, corresponding to a 30.7\% reduction. We also compare the models with and without policy randomization in a zero-shot indoor evaluation, without applying any real-world fine-tuning. The model trained with policy randomization achieves a mean 10-s position RE of 0.0907~m, compared with 0.1098~m for w/o PR. These results show that policy randomization reduces policy-specific overfitting and improves generalization to both unseen policies and real-world conditions.

\begin{table}[t]
    \centering
    \caption{\textbf{Ablation Study III}: Impact of Real-World Fine-Tuning and Comparison with Legolas-FT}
    \label{tab:real_ft_loss_ablation}
    \renewcommand{\arraystretch}{1.15}
    \setlength{\tabcolsep}{3pt}
    \resizebox{\columnwidth}{!}{
    \begin{tabular}{|c|c|c|c|c|c|c|c|}
        \hline
        \multirow{2}{*}{\textbf{Terrain}}
        & \multirow{2}{*}{\textbf{Method}}
        & \textbf{ATE} & \textbf{ATE} & \textbf{ATE}
        & \textbf{RE} & \textbf{RE} & \textbf{RE} \\
        & & \textbf{(pos)} & \textbf{(vel)} & \textbf{(ori)}
        & \textbf{(pos)} & \textbf{(vel)} & \textbf{(ori)} \\
        \hline

        \multirow{4}{*}{\textit{Flat}}
        & \textbf{Proposed}
        & \textbf{0.0882} & \textbf{0.0561} & 0.0352
        & \textbf{0.0538} & 0.0795 & \textbf{0.0172} \\
        \cline{2-8}
        & w/o FT
        & 0.2050 & 0.0567 & 0.0696
        & 0.0573 & \textbf{0.0776} & 0.0246 \\
        \cline{2-8}
        & Full FT
        & 0.1515 & 0.0561 & \textbf{0.0216}
        & 0.0904 & 0.0832 & 0.0181 \\
        \cline{2-8}
        & Legolas-FT
        & 0.2173 & -- & 0.0315
        & 0.0938 & -- & 0.0293 \\
        \hline

        \multirow{4}{*}{\textit{Rough}}
        & \textbf{Proposed}
        & \textbf{0.1656} & 0.1018 & \textbf{0.0221}
        & \textbf{0.0977} & 0.1224 & \textbf{0.0199} \\
        \cline{2-8}
        & w/o FT
        & 0.2435 & 0.1027 & 0.0385
        & 0.1374 & 0.1248 & 0.0233 \\
        \cline{2-8}
        & Full FT
        & 0.2334 & \textbf{0.1012} & 0.0295
        & 0.1044 & \textbf{0.1207} & 0.0226 \\
        \cline{2-8}
        & Legolas-FT
        & 0.6794 & -- & 0.1775
        & 0.2526 & -- & 0.0618 \\
        \hline

        \multirow{4}{*}{\textit{Slippery}}
        & \textbf{Proposed}
        & \textbf{0.1154} & 0.0672 & 0.0246
        & \textbf{0.0550} & \textbf{0.0758} & 0.0128 \\
        \cline{2-8}
        & w/o FT
        & 0.1666 & 0.0683 & \textbf{0.0240}
        & 0.0762 & 0.0775 & \textbf{0.0111} \\
        \cline{2-8}
        & Full FT
        & 0.2248 & \textbf{0.0668} & 0.0347
        & 0.0871 & 0.0760 & 0.0165 \\
        \cline{2-8}
        & Legolas-FT
        & 0.5185 & -- & 0.2290
        & 0.2043 & -- & 0.0700 \\
        \hline

        \multirow{4}{*}{\textit{Soft}}
        & \textbf{Proposed}
        & \textbf{0.1530} & 0.0844 & \textbf{0.0153}
        & 0.0930 & 0.0941 & \textbf{0.0133} \\
        \cline{2-8}
        & w/o FT
        & 0.2043 & \textbf{0.0840} & 0.0406
        & \textbf{0.0920} & \textbf{0.0920} & 0.0162 \\
        \cline{2-8}
        & Full FT
        & 0.2713 & 0.0860 & 0.0226
        & 0.1387 & 0.0984 & 0.0197 \\
        \cline{2-8}
        & Legolas-FT
        & 0.5804 & -- & 0.1729
        & 0.2335 & -- & 0.0614 \\
        \hline
    \end{tabular}
    }
\end{table}

Table~\ref{tab:real_ft_loss_ablation} summarizes the real-world fine-tuning ablation results. We compare the proposed partial fine-tuning strategy, which updates only the GRU and MLP head, with the model without fine-tuning (w/o FT) and full-network fine-tuning (Full FT) using the same real-world dataset and loss functions. Since collecting fine-tuning data from every terrain that a robot may encounter is impractical, we focus on generalization beyond the fine-tuning distribution. Accordingly, although fine-tuning is performed using trajectories collected on outdoor grass and hard ground, all models are evaluated on four indoor terrains. Compared with both w/o FT and Full FT, the proposed method achieves lower position ATE across all terrains, demonstrating the effectiveness of the proposed partial fine-tuning strategy. In particular, compared with Full FT, the proposed partial fine-tuning strategy reduces the position ATE on slippery (48.7\%) and soft (43.6\%) terrains significantly, which differ considerably from the data used for fine-tuning. These results indicate that full-network fine-tuning overfits to the training distribution, whereas partial fine-tuning better preserves simulation-learned representations and generalizes to unseen contact conditions.

Additionally, we compare the proposed estimator with Legolas-FT, which freezes its CNN encoder and masking head and fine-tunes only the pose and variance heads using the same real-world data. Averaged over the four indoor terrains, the proposed estimator reduces the position ATE and RE by 73.8\% and 61.8\%, respectively. Although the proposed estimator uses only 0.2M parameters, whereas Legolas uses 1.4M, it consistently achieves higher estimation accuracy. This result demonstrates the effectiveness of the compact foot-aware architecture.

\section{Conclusion}

We presented TRACE, an end-to-end learned proprioceptive odometry estimator that directly predicts relative motion and body-frame velocity without a filtering backend or manually defined contact and slip thresholds. Foot-aware cross-attention and physics-inspired auxiliary objectives enable the estimator to suppress unreliable leg information and maintain kinematic consistency under challenging contact conditions. Experiments across diverse indoor and outdoor terrains demonstrate improved position accuracy over filtering-based, hybrid, and purely learning-based baselines. Attention analysis shows that the learned attention behaves as a soft measure of foot reliability by emphasizing informative stance legs and downweighting contact feet with large residual motion. The ablation studies further demonstrate that the auxiliary losses improve physical consistency, policy randomization enhances generalization to unseen locomotion policies and real-world conditions, and partial fine-tuning reduces the sim-to-real gap while preserving generalization to diverse terrains. Future work will investigate explicit IMU bias estimation and uncertainty prediction for improved long-horizon accuracy. A further direction is to extend the framework to other legged platforms, such as humanoid robots with different foot geometries and contact dynamics.

\bibliographystyle{IEEEtran}
\bibliography{references}

\end{document}